\documentclass[conference]{IEEEtran}
\IEEEoverridecommandlockouts
\usepackage{cite}
\usepackage{amsmath,amssymb,amsfonts}
\usepackage{algorithmic}
\usepackage{graphicx}
\usepackage{textcomp}
\usepackage{xcolor}
\usepackage{float}
\usepackage{booktabs}
\usepackage{multicol,multirow}
\usepackage[nolist]{acronym}
\usepackage[caption=false]{subfig}
\def\BibTeX{{\rm B\kern-.05em{\sc i\kern-.025em b}\kern-.08em
    T\kern-.1667em\lower.7ex\hbox{E}\kern-.125emX}}

\begin{acronym}[TDMA]
    \acro{AUC-PR}{Area Under the Precision-Recall Curve}
    \acro{AUC-ROC}{Area Under the ROC Curve}
    \acro{LR}{Logistic Regression}
    \acro{SURF}{Speeded Up Robust Features}
    \acro{SVM}{Support Vector Machine}
\end{acronym}

\newif\ifauthors
\authorstrue
    
\begin{document}

\title{Evaluation of Different Annotation Strategies for Deployment of Parking Spaces Classification Systems}

\ifauthors
\author{

    \IEEEauthorblockN{Andre G. Hochuli, Alceu S. Britto Jr.}
	\IEEEauthorblockA{Graduate Program in Informatics\\
	    Pontif\'icia Universidade Cat\'olica do Paran\'a\\
		Curitiba, PR - Brazil\\
		\{aghochuli, alceu\}@ppgia.pucpr.br}
	\and
    \IEEEauthorblockN{Paulo  R. L. de Almeida}
	\IEEEauthorblockA{Department of Informatics\\
	    Universidade Federal do Paran\'a\\
		Curitiba, PR - Brazil\\
		paulorla@ufpr.br}
	
	\and
	\IEEEauthorblockN{Williams B. S. Alves, F\'abio M. C. Cagni}
	\IEEEauthorblockA{
	    Pontif\'icia Universidade Cat\'olica do Paran\'a\\
		Curitiba, PR - Brazil\\
		\{williams.alves, fabio.cagni\}@pucpr.edu.br}
	
}
\else 
\author{
    \IEEEauthorblockN{Anonymous Authors}
    \vspace*{3.5cm}
}
\fi
\maketitle

\begin{abstract}
When using vision-based approaches to classify individual parking spaces between occupied and empty, human experts often need to annotate the locations and label a training set containing images collected in the target parking lot to fine-tune the system. We propose investigating three annotation types (polygons, bounding boxes, and fixed-size squares), providing different data representations of the parking spaces. The rationale is to elucidate the best trade-off between handcraft annotation precision and model performance. We also investigate the number of annotated parking spaces necessary to fine-tune a pre-trained model in the target parking lot. Experiments using the PKLot dataset show that it is possible to fine-tune a model to the target parking lot with less than 1,000 labeled samples, using low precision annotations such as fixed-size squares.

\end{abstract}


\section{Introduction}\label{sec:intro}

Nowadays, urban environments and services need to become innovative. In this vein, taking advantage of machine learning advances, several solutions for \textit{Smart Cities} has been proposed\cite{nosratabadi2019SmartCities}. Likewise, parking spot classification as empty or occupied using machine learning and computer vision techniques. Computer vision-based solutions are widespread since digital cameras may be cheaper and more versatile to monitor parking spots than individual (e.g., ultrasonic) sensors installed in each parking spot.


The recent success of computer vision-based solutions relies on deep learning models. However, in this field of machine learning, most of the approaches demand a vast annotated dataset to learn the model for the desired task\cite{Hochuli2020,Hochuli2020b,zhao2021end,laroca2021efficient,laroca2021towards}. One can argue that deploying a monitoring parking system can also profit from deep learning techniques since we have several public parking lot datasets \cite{almeidaEtAl2015,amatoEtAl2017,nietoEtAl2019}. Furthermore, in the case that still needs annotating and labeling thousands of parking spot samples, this uphill task is necessary only once for training the model. This reasoning is only valid if the environmental conditions are well controlled, i.e., there are no changes in camera position, occlusions, redefinition of parking slots, light changes, etc. Howbeit, usually the environment is dynamic, and the system needs to be updated frequently to fit changes. 


The deployment of a parking lot monitoring system usually requires an initial annotation of the boundary of each parking slot. With this, each parking slot can be cropped and classified to define its status (empty or occupied) \cite{almeidaEtAl2015,AhrnbomAstromNilsson2016,amatoEtAl2017,jensenEtAl2017,nurullayevLee2019,almeidaElAl2020}. The literature demonstrates that even a model trained on many samples must be fine-tuned considering a set of images of the target parking lot. The fine-tuning process reports accuracies close to 99\%, against results that are often less than 95\% without such adaptation to the target domain \cite{almeidaEtAl2015,amatoEtAl2017,vargheseSreelekha2019,almeidaElAl2020}. With this in mind, in this work, the following research questions are considered:


\begin{itemize}
    \item RQ1: Can a relatively cheap approach to demarcate the parking spots positions, such as bounding boxes, provide good results?
    \item RQ2: Considering a pre-trained model, how many labeled samples from the target parking lot are necessary to fine-tune the model?
\end{itemize}


Although the parking spot locations demarcation needs to be performed once at system deployment or maintenance, it is a laborious and time-consuming task. A standard annotation method is to define a polygon bounding each parking spot \cite{almeidaEtAl2015}. Evident that it requires a high level of attention and precision. One way to mitigate this cost is to use low-precision demarcations that are easier to make, such as bounding boxes or fixed-size squares. It is also essential to consider that each annotation type provides different representations, such as encoding contextual information of neighborhoods. So, the performance of models trained with these representations is a matter of discussion.

Considering exposed so far, we also investigate the best trade-off between the amount of labeled data necessary and model performance.

The experimental results show that combining easy to demarcate approaches, such as bounding boxes, and a fine-tuning with less than 1,000 labeled samples (between occupied or empty), can achieve good results. Our findings may lead to the development of vision-based systems that are accurate and that need low human effort to fine-tune the system to the target parking lot.

This work is organized into five sections. Section \ref{sec:state-of-art} presents some important contributions to the subject of parking slot classification. Section \ref{sec:prob_stat} describes the problem statement and proposed protocol to answer our research questions, while Section \ref{sec:experiments} describes the experiments and corresponding results. Finally, Section \ref{sec:conclusion} presents our conclusions and insights for future work.

\section{State-of-the-Art}\label{sec:state-of-art}

When analyzing the state-of-the-art works, we considered no-change and change scenarios. In a \textit{no-change} scenario, the method is trained using images from the same parking lot and camera angle from the test set. In a \textit{change} scenario, the train images must come from a different camera angle or parking lot from the test set. A \textit{change} scenario is often used in the state-of-the-art to test the generalization capabilty of the approaches.

The PKLot, one of the first large-scale parking lot image datasets, was proposed in \cite{almeidaEtAl2015}. The dataset contains 12,417 images collected from two different parking lots (UFPR and PUCPR) and three camera angles (called UFPR04, UFPR05, and PUCPR). There are approximately 700,000 annotated (position and status -- occupied or empty) parking spaces when considering the ground truth. The parking spaces positions are available as polygons and also as rotated rectangles. In the same work, the authors propose the use of texture-based features and ensembles of \ac{SVM} classifiers to classify the parking spaces. The authors reported accuracies over 99\% in no-change scenarios, and ranging from 82\% to 90\% in change scenarios.


Similarly, in \cite{AhrnbomAstromNilsson2016}, \ac{SVM} and \ac{LR} classifiers are trained using the individual color image channels. The PKLot dataset was used in their experiments. The authors reported \ac{AUC-ROC} values over 0.99 for no-change tests, and \ac{AUC-ROC} values ranging from 0.94 to 1.0 under change-scenarios.


A lightweight version of the AlexNet network \cite{krizhevskyEtAl2012} designed to classify the parking spaces was proposed in \cite{amatoEtAl2017}. The authors also proposed the CNRPark-EXT dataset, containing about 160,000 annotated parking spaces, collected from nine different camera angles. The parking spaces were annotated as fixed-size squares. In their experiments, both CNRPark-EXT and PKLot were used. They reported accuracies ranging from 90\% to 98\% in no-change scenarios and 93\% to 98\% in change scenarios.


A lightweight network is also described in \cite{jensenEtAl2017}. The proposed model uses a fixed input size of 40x40 pixels. The authors employed the PKLot dataset and its original test protocol. They reported accuracy over 99\% for no-change scenarios, and ranging from 96\% to 99\% under change scenarios.


In \cite{nietoEtAl2019} the Faster R-CNN \cite{renEtAl2015} was used for classification. Using homographic transformation and perspective correction to put the images into a common plane, the authors fused the classification results of a parking space collected from different perspectives to get the final result. The authors proposed the PLds dataset containing 60,000 annotated bounding boxes of car positions in the same work. The authors reported an \ac{AUC-PR} of 0.92 in no-change scenarios.


A custom network was proposed in \cite{nurullayevLee2019}. The authors employed dilated convolutions in the proposed network to skip pixels in the convolution kernel to increase the network's ability to learn the global context of the images. The authors reported 96\% to 99\% of accuracy under no-change scenarios and 94\% to 98\% under change scenarios.


The authors in \cite{vargheseSreelekha2019} used bag of features combining the \ac{SURF} \cite{bayEtAl2008} descriptor and color information as features, and a \ac{SVM} as classifier.  The authors reported accuracies ranging from 91\% to 100\% under no-change scenarios, and 82\% under a change-scenario.


More recently, \cite{almeidaElAl2020} tested some approaches to deal with concept drifts in the parking spaces classification problem. The authors modeled the problem as a sequence, where the samples of the PKLot dataset are presented in chronological order for classification. Before the classification of the current day, 100 samples of the previous day are given for training. The accuracies reported range from 87\% to 90\% in the modeled change scenario.

Except to \cite{almeidaElAl2020}, the proposed methods in the state-of-the-art train the methods from scratch for the target parking lot (no-change scenario), leading to good results but requiring large amounts of labeled data. For instance, in the original test protocol of the PKLot, the classifiers are trained using about at least 50,000 labeled samples. When using a pre-trained model to classify the target parking lot instances (change scenario), the works in the state-of-the-art can drastically reduce the human effort necessary to deploy such systems, but at the cost of a steep decrease in the accuracy.

\section{Problem Statement}\label{sec:prob_stat}

As discussed in Section \ref{sec:state-of-art}, several approaches in state-of-art rely on thousands of labeled samples to fit a model. The construction of such train datasets can make it impractical for real-world deployments.


In light of this, we postulate that it is possible to reduce the deployment effort by exploiting the bottlenecks of the traditional supervised learning pipeline, which means the manual data annotation of a large training dataset and training models from scratch.

For that, we proposed to use of a pre-trained model. For training this model, we use different demarcation techniques (polygons, bounding boxes, and fixed squares) of the parking spots positions to verify if it is possible to reduce the human effort when deploying such systems. We also check which demarcation strategy leads to the best accuracy when no train data from the target parking lot is given and when a fine-tuning in the target parking lot is performed.

As demonstrated in the state-of-the-art, when using a pre-trained (Section \ref{sec:state-of-art}) model to classify images from a different target parking lot (change scenario), the accuracies may drop to values below 90\%. To mitigate this, we evaluate the trade-off between the number of labeled samples collected in the target parking lot necessary to fine-tune the model and the human effort necessary to manually label parking lot samples.

\subsection{Network Architecture}
 
\begin{figure*}[htpb]
      \centering
      \includegraphics[width=0.85\textwidth]{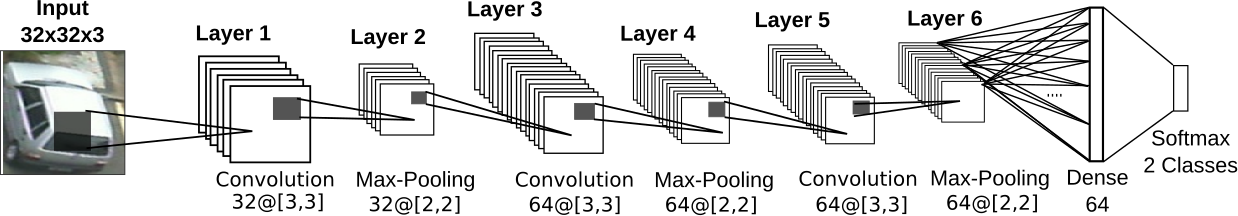}
      \caption{The 3-convolutional layers architecture used to classify the parking spots.}
      \label{fig:cnn_arch_pklot}
    \end{figure*}

We defined a CNN architecture composed of three convolutional layers followed by max-polling to classify parking spots. In the end, a dense layer concatenate all features and discriminate them into two classes: i) empty or ii) occupied. The model overview is depicted in Figure \ref{fig:cnn_arch_pklot}. Its architecture was defined empirically and achieved feasible results. A deep discussion of the performance is presented in Section \ref{sec:experiments}. 
    
    
The convolutional and dense layers have learnable parameters optimized during training. Except for the last layer in the network, after each learnable layer, we apply ReLU non-linearity. The final layer uses the softmax activation function.


Training is performed with the Stochastic Gradient Descent (SGD) using back-propagation with mini-batches of 32 instances, a momentum factor of 0.9, and a weight decay of $5 \times 10^{-4}$. The learning rate is set to $10^{-3}$, in the beginning, to allow the weights to quickly fit the long ravines in the weight space, after which it is reduced over time (until  $5 \times 10^{-4}$) to make the weights fit the sharp curvatures. The network makes use of the well-known cross-entropy loss function.


In this work, regularization was implemented through early-stopping, which prevents overfitting by interrupting the training procedure once the network's performance on a validation set deteriorates\footnote{All trained classifiers, data, and sample codes are available for research purposes at \textbf{https://github.com/andrehochuli/pklot-eval-annotations}}. 


\subsection{Parking Lot Dataset}

\begin{figure}[]
  \centering    
  \subfloat[UFPR04]{\includegraphics[width=0.229\textwidth]{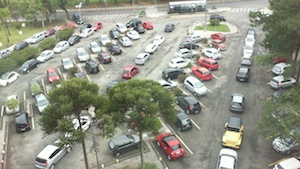}}\quad
  \subfloat[UFPR05]{\includegraphics[width=0.229\textwidth]{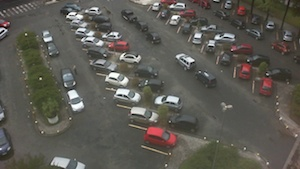}}\quad
  
  \subfloat[PUCPR]{\includegraphics[width=0.25\textwidth]{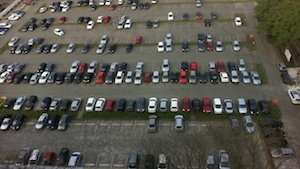}}
  \caption{The PKLOT dataset comprises three scenarios named a) UFPR04, b) UFPR05 and c) PUCPR.}
  \label{fig:overview_pklot}
\end{figure}

In this work, we used the PKLot dataset\cite{almeidaEtAl2015}, which comprises three different scenarios (two camera angles for the UFPR parking lot and one camera angle for the PUCPR parking lot). For each scenario, the images were collected for approximately thirty days, with an interval of five minutes between camera shots. The dataset provides robust scenarios containing various conditions, including variations of camera positions, background, weather, and occlusions. It also offers annotations for the bounding polygons of each parking spot and their labels (occupied or empty) over time. A PKLot overview is depicted in Figure \ref{fig:overview_pklot}.

\begin{figure}[htpb]
  \centering    
  \subfloat[Polygon\label{subfig:exPoly}]{\includegraphics[width=0.145\textwidth]{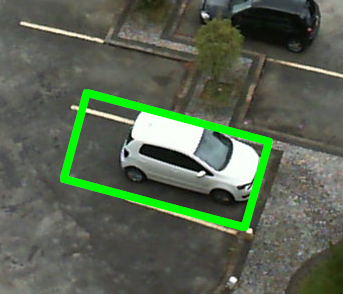}}\quad
  \subfloat[Bounding Box\label{subfig:exBBox}]{\includegraphics[width=0.145\textwidth]{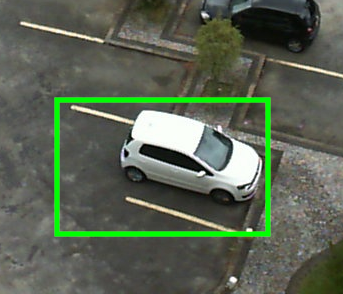}}\quad
  \subfloat[Fixed Square\label{subfig:fixed}]{\includegraphics[width=0.145\textwidth]{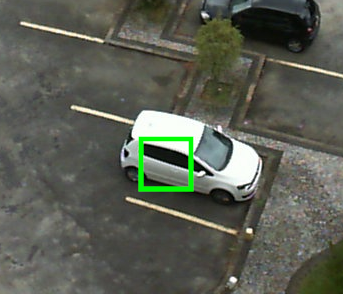}}
  \caption{Types of parking spots location demarcations.}
\end{figure}

These annotated boundaries are important since, through the polygons, we can compute the bounding boxes and the fixed-size squares. A polygon annotation encompasses the parking space location (see Figure \ref{subfig:exPoly}). From that location, a bounding box is defined as a non-rotated rectangle that must encompass the entire parking space (see Figure \ref{subfig:exBBox}). A fixed-size square is defined as a square of side $N$ (the same value $N$ is used for all parking spaces), where the square is positioned approximately in the center of the parking spot (see Figure \ref{subfig:fixed}).

The rationale here is that those annotation types represent the problem differently since they address distinct levels of contextual information. In addition, the level of human attention and time-consuming for each annotation type is different. Hence, the experiments should elucidate the best trade-off between annotation quality and performance. 


\subsection{Evaluation Protocol}

Since the images are taken at an interval of five minutes, the evaluation protocol follows that described by Almeida et al.\cite{almeidaEtAl2015}. For each scenario (UFPR04, UFPR05, and PUCPR), we select the first 50\% of days for the training set and the remaining 50\% of days as the testing set. This strategy avoids that a vehicle parked for a long time or an empty spot with slight variation in light belongs to both training and test sets simultaneously. The number of parking spots on each set is shown in Table \ref{tab:pklot_protocol}. To avoid unbalanced training, we randomly drop images from the class with most samples to match the minority class in the training set. We take 20\% of samples from the training set for validation.

\begin{table}[!htpb]
\caption{Evaluation protocol defined for the PKLot dataset\cite{almeidaEtAl2015}.}
\label{tab:pklot_protocol}
\begin{tabular}{@{}ccccccc@{}}
\toprule
\multirow{2}{*}{\textbf{Classes}} & \multicolumn{2}{c}{\textbf{UFPR04}} & \multicolumn{2}{c}{\textbf{UFPR05}} & \multicolumn{2}{c}{\textbf{PUCPR}} \\
 & \textbf{Train} & \textbf{Test} & \textbf{Train} & \textbf{Test} & \textbf{Train} & \textbf{Test} \\ \cmidrule(l){1-7} 
\textbf{OCCUPIED} & 24554 & 21571 & 46994 & 50432 & 93804 & 100425 \\
\textbf{EMPTY} & 31956 & 27764 & 36275 & 32084 & 118689 & 111351 \\ \cmidrule(l){1-7} 
\textbf{TOTAL} & 56510 & 49335 & 83269 & 82516 & 212493 & 211776 \\ \bottomrule
\end{tabular}

\end{table}


It is worth mentioning that there are other important datasets, such as CNRPark-Ext \cite{amatoEtAl2017}, and PLds \cite{nietoEtAl2019}. However, they contain only bounding boxes or fixed-size squares annotations, in which a manual translation to polygons would be necessary to encompass the exact parking spaces in this work.

\section{Experiments}\label{sec:experiments}

\subsection{Evaluation of Annotation Effort}

We created a simple interactive tool to demarcate the parking spots in an image using a polygon, a bounding box, and a fixed-size square to give us insight into the time necessary for a human to demarcate the parking spaces. To annotate a polygon, the user needs to define four corners of each parking spot. For a bounding box, the user provides the upper left and bottom right points to create a box encompassing the parking spot. Finally, by giving one point that is approximately at the center of the image, a fixed-size square of 32x32 pixels is defined.

\begin{table}[htpb]
\centering
\caption{Average time necessary to demarcate a parking spot.}
\label{table:averageTimeDemarc}
\begin{tabular}{lr}
\textbf{Type} & \textbf{Average time per spot} \\\hline
Fixed-Size   & 0.9 seconds           \\
Bounding Box & 2.7 seconds           \\
Polygon      & 3.9 seconds          \\\hline
\end{tabular}
\end{table}

The average results from our team are presented in Table \ref{table:averageTimeDemarc}, where each person demarcated the positions of on average 42 parking spots using each approach (fixed-size rectangles, bounding boxes, and polygons). As one can observe, our intuition that fixed-size squares and bounding box annotations require less manual effort to demarcate is supported by this empirical test.

\subsection{On the Annotation Quality and Cross Domain Analysis}\label{sec:exp1_crossdomain}

To properly answer the research question RQ1 (Section \ref{sec:intro}), we trained the models using the protocol and annotations methods discussed in Section \ref{sec:prob_stat}. This experiment aims to figure out the impact of annotation quality in the classification of parking spots. 
    
Moreover, we extend to cross-domain analysis, which means evaluating the performance of a model trained in one parking lot (source) and testing on unseen spots for a different parking lot or camera angle (target). This can be considered a \textit{change} scenario, as discussed in Section \ref{sec:state-of-art}. It corroborates to elucidate whether there is a necessity to adjust a pre-trained model on a new target.
    
The results depicted on Table \ref{tab:annotation_results} allows us to draw some conclusions. As expected, the performance on the \textit{non-change} scenarios achieved performances close to 99\% independently of the annotation type since we have a robust training set with several annotated samples.
    
The polygon annotation reports better rates for cross-domain (\textit{change}) scenarios, while the fixed-size square performed worst. The rationale is that the fixed-size squares do not encode the neighborhood's contextual information, struggling with the generalization. The importance of contextual information in machine learning was well-demonstrated in \cite{divala2009ContextualInfo}. As also demonstrated by the state-of-the-art (Section \ref{sec:state-of-art}), there are steep accuracy drops for all annotation types when considering the cross-domain results in Table \ref{tab:annotation_results}.
    
    
The bounding-box annotation plays the best trade-off between annotation effort and accuracy since it achieved accuracies closest to the polygon, with a considerably reduced effort to annotate the spots.

\begin{table}[!hbtp]
	\caption{Accuracies (\%) of models trained with different annotation types of PKLot dataset and Tested in a cross-domain strategy}
	\label{tab:annotation_results}
	\resizebox{0.48\textwidth}{!}{%
		\begin{tabular}{ccccccc}
			\multicolumn{7}{c}{\textbf{Polygon}}                                                             \\\hline                                                                                
			\multirow{3}{*}{\textbf{Models}} 
			& \multicolumn{2}{c}{\textbf{UFPR04 Test Set}} & \multicolumn{2}{c}{\textbf{UFPR05 Test Set}} & \multicolumn{2}{c}{\textbf{PUCPR Test Set}} \\
			& \textbf{W/O Aug}      & \textbf{W/ Aug}      & \textbf{W/O Aug}      & \textbf{W/ Aug}      & \textbf{W/O Aug}      & \textbf{W/ Aug}     \\ \hline
			\textbf{UFPR04}                      & 98,5        & 98,2       & 92,1        & 89,1       & 98,4        & 97,4      \\
			\textbf{UFPR05}                      & 91,4        & 89,4       & 99,5        & 99,5       & 91,8        & 90,7      \\
			\textbf{PUCPR}                       & 98,9        & 98,0        & 92,0         & 90,4       & 99,8        & 99,8      \\ \hline
			&              &             &              &             &              &            \\
			
			\multicolumn{7}{c}{\textbf{Bounding Box}}                                                          \\\hline
			
			\multirow{3}{*}{\textbf{Models}} 
			& \multicolumn{2}{c}{\textbf{UFPR04 Test Set}} & \multicolumn{2}{c}{\textbf{UFPR05 Test Set}} & \multicolumn{2}{c}{\textbf{PUCPR Test Set}} \\
			& \textbf{W/O Aug}      & \textbf{W/ Aug}      & \textbf{W/O Aug}      & \textbf{W/ Aug}      & \textbf{W/O Aug}      & \textbf{W/ Aug}     \\ \hline
			\textbf{UFPR04}                      & 98,6        & 99,1       & 77,7        & 74,6       & 96,7        & 96,7      \\
			\textbf{UFPR05}                      & 92,1        & 92,5       & 99,5        & 99,6       & 87,7        & 89,0       \\
			\textbf{PUCPR}                       & 96,7        & 96,7       & 89,2        & 90,4       & 99,8        & 99,8      \\ \hline
			&              &             &              &             &              &            \\
			
			\multicolumn{7}{c}{\textbf{Fixed Square}}                                                          \\ \hline                 
			\multirow{3}{*}{\textbf{Models}} 
			& \multicolumn{2}{c}{\textbf{UFPR04 Test Set}} & \multicolumn{2}{c}{\textbf{UFPR05 Test Set}} & \multicolumn{2}{c}{\textbf{PUCPR Test Set}} \\
			& \textbf{W/O Aug}      & \textbf{W/ Aug}      & \textbf{W/O Aug}      & \textbf{W/ Aug}      & \textbf{W/O Aug}      & \textbf{W/ Aug}     \\ \hline
			\textbf{UFPR04}                      & 97,4        & 93,7       & 60,5        & 71,0        & 89,2        & 71,0       \\
			\textbf{UFPR05}                      & 83,4        & 89,8       & 99,3        & 98,9       & 87,0         & 91,3      \\
			\textbf{PUCPR}                       & 94,7        & 95,4       & 91,8        & 93,2       & 99,7        & 99,8      \\ \hline
			\multicolumn{7}{l}{$^{*}$With (W/) or Without (W/O) Data Augmentation (Aug)}  
		\end{tabular}%
	}
\end{table}

We also tested the use of some classic data augmentation techniques (flip, saturation, contrast, shift, and zoom). As one can observe in Table \ref{tab:annotation_results}, the use of data augmentation did not contribute to a better performance in most scenarios. The large number of train samples under various circumstances may explain this. The data augmentation could not provide any new relevant information to the network that was not present in the training set.
    
Notice that when considering cross-domain performances, the model trained using the UFPR04 subset of the PKLot has the worst results. As we can observe in Table \ref{tab:pklot_protocol}, this parking lot provided the lowest number of samples resulting in a poor model generalization. On the other hand, PUCPR has the highest overall rates in most scenarios since this subset provides variations in slot sizes due to camera angle and number of parking slots, giving different types of occlusion, shadows, brightness conditions, and vehicle types.

\subsection{Fine-Tuning and Catastrophic Forgetting}

In Section \ref{sec:exp1_crossdomain}, the cross-domain performance achieved interesting results. In several applications, the use of fine-tunning over a small set of samples from the target scenario improves the recognition rates considerably \cite{WANG2018}.

    
Addressing the question RQ2, we are interested in a three-fold investigation of the fine-tuning approach:  i) the performance on the target dataset, ii) the performance of source data (catastrophic forgetting), and iii) the trade-off of the number of samples versus accuracy rate.
    
The experiment protocol is defined as follows: We defined different sets of samples randomly taken from the train set of the target dataset. Each set is composed of $N \in [50, 100, 200, 500, 1K, 2K, 5K, 10K, 20K, 50K]$ images. The weights were of a pre-trained model are fine-tuned over 5-epochs for each different value of $N$. The presented results are an average of 10 runs.

\begin{figure}[!htbp]
    \centering
    \includegraphics[width=0.45\textwidth]{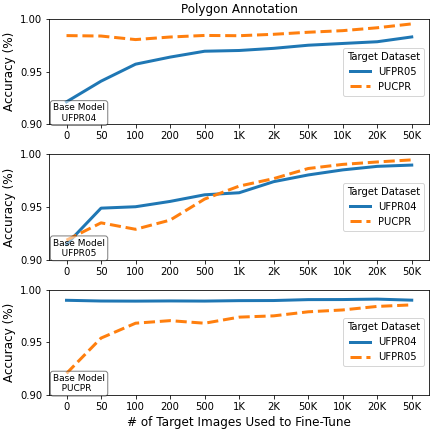}
    \caption{Cross-domain performance using polygon annotation.}
    \label{fig:poly-finnetunned}
\end{figure}

\begin{figure}[!htbp]
    \centering
    \includegraphics[width=0.45\textwidth]{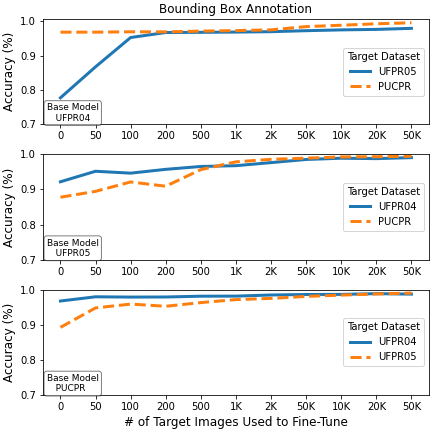}
    \caption{Cross-domain performance using bounding box annotation.}
    \label{fig:bbox-finnetunned}
\end{figure}

\begin{figure}[!htbp]
    \centering
    \includegraphics[width=0.45\textwidth]{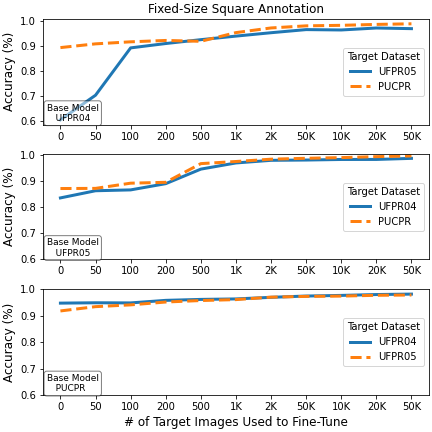}
    \caption{Cross-domain performance using fixed-size square annotation.}
    \label{fig:fixedsquare-finnetunned}
\end{figure}

The results for polygons, bounding boxes and fixed-size squares are depicted in Figures \ref{fig:poly-finnetunned}, \ref{fig:bbox-finnetunned}, and \ref{fig:fixedsquare-finnetunned}, respectively. In the Figures, the base model stands for the original dataset used for the model's training.
    
The conclusion is straightforward. With just a small set composed of 1,000 samples, which is feasible to a human annotate, the models can reach up to 97\% performance.
In addition, the fine-tunning step can lead to good results even when using fixed squares, which is fast to annotate according to the experiment reported by Table \ref{table:averageTimeDemarc}. Besides, there is a gain in computational cost since fine-tuning is cheaper than training from scratch.

Towards a model generalization, clarifying whether a catastrophic forgetting on the source parking lot occurs is essential. Since the fixed-size square annotation provides the best trade-off, we evaluate the fine-tuned model testing it on the test data of the same camera and parking lot where the model was pre-trained. In Figure \ref{fig:fixedsquare-finnetunned} we can note that the base model UFPR04 has a significant forgetting with a small set of samples, which should be the amount of data reasonable to annotate in a real-world environment. The model may overfit the target domain since the fixed-size square has less contextual information. The catastrophic forgetting may be a problem when, for instance, the goal is to create a generalist model able to classify images from different cameras. This is an important issue that will be explored in future research.
    
\begin{figure}[!htbp]
    \centering
    
    \includegraphics[width=0.45\textwidth]{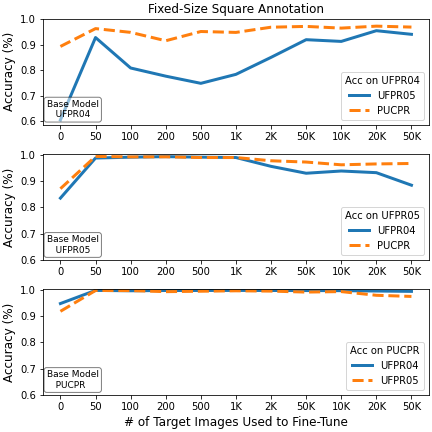}
    \caption{Evaluation of the accuracy on the source data using fixed-size square annotations.}
    \label{fig:fixedsquare-loss}
\end{figure}    
 
\section{Conclusion}\label{sec:conclusion}
   
We evaluated three annotation schemes for parking lots (polygons, bounding boxes, and fixed-size squares), considering the best trade-off between the annotation pixel precision and the performance of deep models trained on them. Moreover, we investigated the number of annotated parking spots necessary to fine-tune a pre-trained model for a new target parking lot. 

A robust experimental protocol based on the well-known PKLot dataset has shown that a bounding box annotation schema can lead to similar results compared to polygons, with the advantage of being less expensive for humans to demarcate.

Towards a real-world deployment scenario, cross-domain experiments performed with three distinct scenarios available in the PKLot dataset have shown that with a small training set (1,000 samples), we may fine-tune a pre-trained CNN model satisfactorily to a new target domain. This tuning is possible even when using fixed-size squares to demarcate the parking spot positions. The fixed-size squares are the cheapest annotation type for parking spot positions. The cross-domain experiments also demonstrated that the well-known catastrophic forgetting problem plays a role in adjusting the models to a target parking lot, making the model lose accuracy in its original training parking lot.

Further work is necessary to investigate strategies to mitigate the catastrophic forgetting problem, making it possible to learn continually in the case of parking lot applications. Solving such a problem is very important to make parking lot classification-based systems scalable.

\section*{Acknowledgment}
The authors would like to thank Conselho Nacional de Desenvolvimento Científico e Tecnológico (CNPq, Brazil, grant 306684/2018-7), and Coordenação de Aperfeiçoamento de Pessoal de Nível Superior (CAPES, Brazil)

\bibliographystyle{IEEEtran}
\bibliography{IEEEabrv,bibliography}

\begin{thebibliography}{10}
\providecommand{\url}[1]{#1}
\csname url@samestyle\endcsname
\providecommand{\newblock}{\relax}
\providecommand{\bibinfo}[2]{#2}
\providecommand{\BIBentrySTDinterwordspacing}{\spaceskip=0pt\relax}
\providecommand{\BIBentryALTinterwordstretchfactor}{4}
\providecommand{\BIBentryALTinterwordspacing}{\spaceskip=\fontdimen2\font plus
\BIBentryALTinterwordstretchfactor\fontdimen3\font minus
  \fontdimen4\font\relax}
\providecommand{\BIBforeignlanguage}[2]{{%
\expandafter\ifx\csname l@#1\endcsname\relax
\typeout{** WARNING: IEEEtran.bst: No hyphenation pattern has been}%
\typeout{** loaded for the language `#1'. Using the pattern for}%
\typeout{** the default language instead.}%
\else
\language=\csname l@#1\endcsname
\fi
#2}}
\providecommand{\BIBdecl}{\relax}
\BIBdecl

\bibitem{nosratabadi2019SmartCities}
S.~Nosratabadi, A.~Mosavi, R.~Keivani, S.~Ardabili, and F.~Aram, ``State of the
  art survey of deep learning and machine learning models for smart cities and
  urban sustainability,'' in \emph{International Conference on Global Research
  and Education}.\hskip 1em plus 0.5em minus 0.4em\relax Springer, 2019, pp.
  228--238.

\bibitem{Hochuli2020}
A.~G. {Hochuli}, A.~S. {Britto}, J.~P. {Barddal}, R.~{Sabourin}, and L.~E.~S.
  {Oliveira}, ``An end-to-end approach for recognition of modern and historical
  handwritten numeral strings,'' in \emph{2020 International Joint Conference
  on Neural Networks (IJCNN)}, 2020, pp. 1--8.

\bibitem{Hochuli2020b}
\BIBentryALTinterwordspacing
A.~G. Hochuli, A.~S. {Britto Jr}, D.~A. Saji, J.~M. Saavedra, R.~Sabourin, and
  L.~S. Oliveira, ``A comprehensive comparison of end-to-end approaches for
  handwritten digit string recognition,'' \emph{Expert Systems with
  Applications}, vol. 165, p. 114196, 2021. [Online]. Available:
  \url{http://www.sciencedirect.com/science/article/pii/S0957417420309271}
\BIBentrySTDinterwordspacing

\bibitem{zhao2021end}
M.~Zhao, A.~G. Hochuli, and A.~Cheddad, ``End-to-end approach for recognition
  of historical digit strings,'' in \emph{International Conference on Document
  Analysis and Recognition}.\hskip 1em plus 0.5em minus 0.4em\relax Springer,
  2021, pp. 595--609.

\bibitem{laroca2021efficient}
R.~{Laroca}, L.~A. {Zanlorensi}, G.~R. {Gon{\c{c}}alves}, E.~{Todt}, W.~R.
  {Schwartz}, and D.~{Menotti}, ``An efficient and layout-independent automatic
  license plate recognition system based on the {YOLO} detector,'' \emph{IET
  Intelligent Transport Systems}, vol.~15, no.~4, pp. 483--503, 2021.

\bibitem{laroca2021towards}
R.~{Laroca}, A.~B. {Araujo}, L.~A. {Zanlorensi}, E.~C. {De Almeida}, and
  D.~{Menotti}, ``Towards image-based automatic meter reading in unconstrained
  scenarios: A robust and efficient approach,'' \emph{IEEE Access}, vol.~9, pp.
  67\,569--67\,584, 2021.

\bibitem{almeidaEtAl2015}
P.~R. Almeida, L.~S. Oliveira, A.~S. Britto~Jr, E.~J. Silva~Jr, and A.~L.
  Koerich, ``Pklot--a robust dataset for parking lot classification,''
  \emph{Expert Systems with Applications}, vol.~42, no.~11, pp. 4937--4949,
  2015.

\bibitem{amatoEtAl2017}
G.~Amato, F.~Carrara, F.~Falchi, C.~Gennaro, C.~Meghini, and C.~Vairo, ``Deep
  learning for decentralized parking lot occupancy detection,'' \emph{Expert
  Systems with Applications}, vol.~72, pp. 327--334, 2017.

\bibitem{nietoEtAl2019}
R.~{Mart{\'i}n Nieto}, {\'A}.~{García-Mart{\'i}n}, A.~G. {Hauptmann}, and
  J.~M. {Mart{\'i}nez}, ``Automatic vacant parking places management system
  using multicamera vehicle detection,'' \emph{IEEE Transactions on Intelligent
  Transportation Systems}, vol.~20, no.~3, pp. 1069--1080, March 2019.

\bibitem{AhrnbomAstromNilsson2016}
M.~Ahrnbom, K.~Astrom, and M.~Nilsson, ``Fast classification of empty and
  occupied parking spaces using integral channel features,'' in
  \emph{Proceedings of the IEEE Conference on Computer Vision and Pattern
  Recognition Workshops}, vol. 2016.\hskip 1em plus 0.5em minus 0.4em\relax
  IEEE, 2016, pp. 1609--1615.

\bibitem{jensenEtAl2017}
T.~H. Jensen, H.~T. Schmidt, N.~D. Bodin, K.~Nasrollahi, and T.~B. Moeslund,
  ``Parking space occupancy verification-improving robustness using a
  convolutional neural network,'' in \emph{International Conference on Computer
  Vision Theory and Applications}, vol.~6.\hskip 1em plus 0.5em minus
  0.4em\relax SCITEPRESS, 2017, pp. 311--318.

\bibitem{nurullayevLee2019}
S.~Nurullayev and S.-W. Lee, ``Generalized parking occupancy analysis based on
  dilated convolutional neural network,'' \emph{Sensors}, vol.~19, no.~2, p.
  277, 2019.

\bibitem{almeidaElAl2020}
P.~R. L.~d. Almeida, L.~S. Oliveira, A.~d. Souza~Britto, and J.~Paul~Barddal,
  ``Naïve approaches to deal with concept drifts,'' in \emph{2020 IEEE
  International Conference on Systems, Man, and Cybernetics (SMC)}, 2020, pp.
  1052--1059.

\bibitem{vargheseSreelekha2019}
A.~{Varghese} and G.~{Sreelekha}, ``An efficient algorithm for detection of
  vacant spaces in delimited and non-delimited parking lots,'' \emph{IEEE
  Transactions on Intelligent Transportation Systems}, vol.~21, no.~10, pp.
  4052--4062, 2020.

\bibitem{krizhevskyEtAl2012}
A.~Krizhevsky, I.~Sutskever, and G.~E. Hinton, ``Imagenet classification with
  deep convolutional neural networks,'' in \emph{Advances in neural information
  processing systems}, 2012, pp. 1097--1105.

\bibitem{renEtAl2015}
S.~Ren, K.~He, R.~Girshick, and J.~Sun, ``Faster r-cnn: Towards real-time
  object detection with region proposal networks,'' in \emph{Advances in neural
  information processing systems}, 2015, pp. 91--99.

\bibitem{bayEtAl2008}
H.~Bay, A.~Ess, T.~Tuytelaars, and L.~Van~Gool, ``Speeded-up robust features
  (surf),'' \emph{Computer vision and image understanding}, vol. 110, no.~3,
  pp. 346--359, 2008.

\bibitem{divala2009ContextualInfo}
S.~K. Divvala, D.~Hoiem, J.~H. Hays, A.~A. Efros, and M.~Hebert, ``An empirical
  study of context in object detection,'' in \emph{2009 IEEE Conference on
  Computer Vision and Pattern Recognition}, 2009, pp. 1271--1278.

\bibitem{WANG2018}
\BIBentryALTinterwordspacing
M.~Wang and W.~Deng, ``Deep visual domain adaptation: A survey,''
  \emph{Neurocomputing}, vol. 312, pp. 135--153, 2018. [Online]. Available:
  \url{https://www.sciencedirect.com/science/article/pii/S0925231218306684}
\BIBentrySTDinterwordspacing

\end{thebibliography}

\end{document}